\documentclass[10pt,twocolumn,letterpaper]{article}

\usepackage[pagenumbers]{cvpr} 

\usepackage{graphicx}
\usepackage{amsmath}
\usepackage{amssymb}
\usepackage{booktabs}
\usepackage{multirow}
\usepackage{makecell}
\usepackage{colortbl}
\usepackage[ruled]{algorithm2e}
\newcommand{\tabincell}[2]{\begin{tabular}{@{}#1@{}}#2\end{tabular}}

\usepackage[pagebackref,breaklinks,colorlinks]{hyperref}

\usepackage[capitalize]{cleveref}
\crefname{section}{Sec.}{Secs.}
\Crefname{section}{Section}{Sections}
\Crefname{table}{Table}{Tables}
\crefname{table}{Tab.}{Tabs.}

\def\cvprPaperID{6416} 
\def\confName{CVPR}
\def\confYear{2022}

\begin{document}

\title{ Revisiting Open World Object Detection}

\author{Xiaowei Zhao\\
Beihang University\\

{\tt\small xiaoweizhao@buaa.edu.cn}
\and
Xianglong Liu\\
Beihang University\\
{\tt\small xlliu@buaa.edu.cn}

\and
Yifan Shen\\
Beihang University\\
{\tt\small shenyf@buaa.edu.cn}

\and
Yixuan Qiao\\
Beihang University\\
{\tt\small qiaoyixuan@buaa.edu.cn}
\and
Yuqing Ma$^*$\\
Beihang University\\
{\tt\small mayuqing@buaa.edu.cn}
\and
Duorui Wang\\
Beihang University\\
{\tt\small by2106125@buaa.edu.cn}
}
\maketitle

\begin{abstract}
{
Open World Object Detection (OWOD), simulating the real dynamic world where knowledge grows continuously, attempts to detect both known and unknown classes and incrementally learn the identified unknown ones. We find that although the only previous OWOD work constructively puts forward to the OWOD definition, the experimental settings are unreasonable with the illogical benchmark, confusing metric calculation, and inappropriate method. In this paper, we rethink the OWOD experimental setting and propose five fundamental benchmark principles to guide the OWOD benchmark construction. Moreover, we design two fair evaluation protocols specific to the OWOD problem, filling the void of evaluating from the perspective of unknown classes. Furthermore, we introduce a novel and effective OWOD framework containing an auxiliary Proposal ADvisor (PAD) and a Class-specific Expelling Classifier (CEC). The non-parametric PAD could assist the RPN in identifying accurate unknown proposals without supervision, while CEC calibrates the over-confident activation boundary and filters out confusing predictions through a class-specific expelling function. Comprehensive experiments conducted on our fair benchmark demonstrate that our method outperforms other state-of-the-art object detection approaches in terms of both existing and our new metrics.\footnote{Our benchmark and code are available at https://github.com/RE-OWOD/RE-OWOD. }
}    

\end{abstract}

\section{Introduction}
\label{sec:intro}

\begin{figure}[t]
    \centering
    \includegraphics[scale = 0.31]{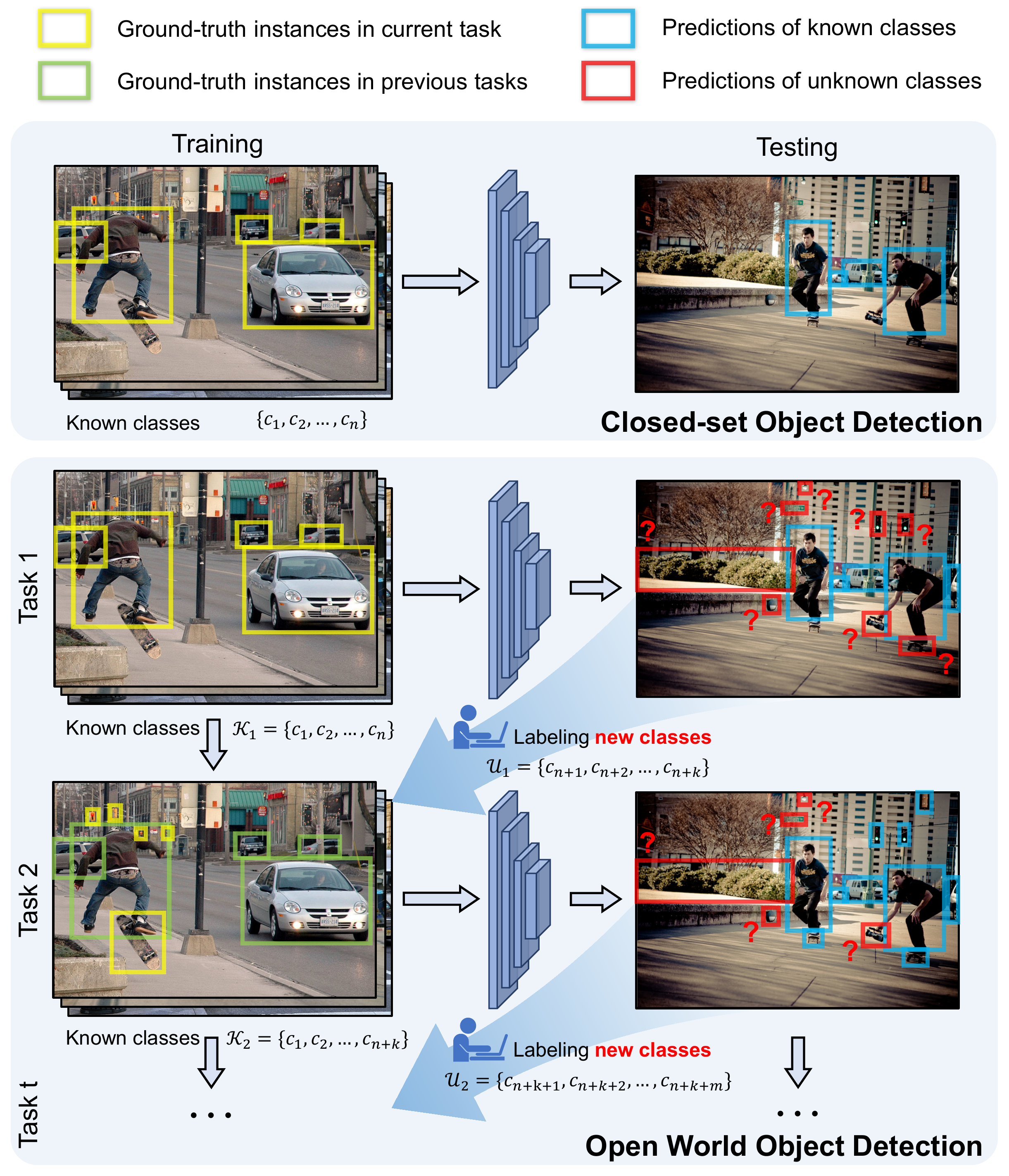}
    \caption{In closed-set object detection, classes of the training set and the testing set are the same, and objects of other classes are simply ignored or mistakenly classified into known classes. While in open world object detection, the unknown instances of the current task will be detected as \emph{unknown}. Then human annotators can gradually assign some labeling information, and the model should incrementally improve its detection performance for those newly-annotated known classes without forgetting.}
    \label{img:intro}
    \vspace{-1em}
\end{figure}


{Deep learning models significantly promote the performance of object detection networks \cite{cao2020d2det,zhu2021semantic}, benefiting a wide range of applications, such as VQA\cite{jing2020overcoming,niu2021counterfactual}, visual reasoning\cite{ge2021peek,hong2021transformation}, and automatic driving \cite{reading2021categorical}.}
{However, the previous learning-based object detection relies on a static closed-set assumption. As shown in Figure \ref{img:intro}, in the closed-set object detection, only known (labeled) classes need to be detected and are used to evaluate the detection performance. Instances of unknown (unlabeled) classes are simply ignored or mistakenly classified as existing known classes. On the contrary, in real life, there are many unknown classes, which humans can easily distinguish from the classes they have already known and are capable of rapidly recognizing them once the labels are assigned. Therefore, to bridge the intelligence between humans and machines and promote the practical application in the open world, it is of great significance to study the open world object detection problem.}
There have been some methodologies for open {world} recognition \cite{xu2019open,bendale2015towards}. However, the complex and challenging {open world object detection} (OWOD) problem is rarely explored. \cite{joseph2021towards} proposed the first OWOD experimental setup and a baseline method ORE model. As shown in Figure \ref{img:intro}, {the OWOD task contains multiple incremental tasks. In each task, the OWOD model only trained by known classes should correctly detect the known classes and recognize the unknown classes as ``unknown" in the test phase. 
Then human annotators can gradually assign labels to the classes of interest, and the detection model in the next task should incrementally learn these classes with the newly-added annotations. 
}

{We find that although \cite{joseph2021towards} defined this practical open world object detection problem, they did not make a rigorous inspection of their benchmark and approach, which are severely against the OWOD definition. There are three unreasonable aspects: 1) illogical benchmark construction,  2) confusing metric calculation, and 3) inappropriate method implementation. More specific explanations will be placed in the following sections and the appendix. 
}

{Therefore, in this paper, we revisit the OWOD task and rethink the experimental settings. We summarize five basic benchmark principles to guide the OWOD benchmark construction, and construct a sound and fair benchmark for the OWOD problem. Moreover, we further introduce two fair metrics, Unknown Detection Recall (UDR) and Unknown Detection Precision (UDP), which are specific to the OWOD task and evaluate the detection performance from the perspective of unknown classes. Our metrics are the first evaluation metrics stressing the importance of distinction from the background, which is a key challenge, especially in detection tasks without supervision. 

}

{Furthermore, we propose a simple and effective OWOD framework with an auxiliary Proposal ADvisor (PAD) module and the Class-specific Expelling Classifier (CEC), to overcome the difficult distinctions between the unknown classes and the background or known classes. The non-parametric PAD could assist the RPN in confirming accurate unknown proposals without supervision, and further guide the learning of RPN to distinguish unknown proposals from the background. While CEC calibrates the over-confident activation boundary and filters out confusing predictions through a class-specific expelling function, avoiding the detection model overconfidence in classifying unknown instances into known classes.}

We summarize our contributions as follows:

{(1) We revisit the Open World Object Detection problem, and summarize the basic benchmark principles to guide the OWOD benchmark construction. }

{(2) Two fair evaluation metrics specific to the OWOD task are proposed, filling the void of evaluating from the perspective of unknown classes.}

{(3) We propose a simple yet effective OWOD framework including an auxiliary Proposal ADvisor (PAD) and a Class-specific Expelling Classifier (CEC), which can assist RPN in identifying unknown proposals and expelling confusing predictions for each known class.}



{(4) Extensive experiments conducted on our fair benchmark demonstrate that our framework consistently yields superior results according to existing and our new metrics. }







\section{Related Work}
Recently, the open set learning problem, which considers the incompleteness of the knowledge in training set during learning, has aroused extensive attention. Open world learning extends the open-set setting with incremental learning, which is more flexible and practical. In this section, we will respectively introduce the recognition and detection task in open-set and open-world settings.

\subsection{Open Set and Open World Recognition}
Scheirer et al. \cite{scheirer2012toward} first formalize the definition of open-set recognition (OSR) as a constrained optimization problem. The follow-up works \cite{scheirer2014probability,jain2014multi,yue2021counterfactual} extend the OSR to multiclass classification combined with the extreme value theory. Bendale et al. \cite{bendale2015towards} first present and define the Open World Recognition (OWR) problem, introduce the Nearest Non-Outlier algorithm, and provide a protocol to evaluate OWR system. Yue et al. \cite{yue2021counterfactual} indicate that OSR needs counterfactual faithful generations and applies the Consistency Rule to perform unseen/seen binary classification.

\subsection{Open Set and Open World Object Detection}
Miller et al. \cite{miller2018dropout} investigate the utility of Dropout Sampling for object detection to extract label uncertainty and use the uncertainty to increase object detection performance under the open-set conditions. Dhamija et al. \cite{dhamija2020overlooked} formally discuss the ability of algorithms to avoid false detection and provide an evaluation metric to analyze their performance on differentiating between close-set and open-set. Joseph et al. \cite{joseph2021towards} extend open set object detection task to open world, they introduce an OWOD model ORE based on contrastive clustering and energy-based unknown detection.

\section{Rethinking Open World Object Detection}
{Open World Object Detection (OWOD) simulates the object detection task in the real dynamic world and bridges the gap between human and machine intelligence, attracting widespread attention.} {However, although \cite{joseph2021towards} constructively defined the OWOD task, their factual experimental setting was unreasonable and incomplete, severely violating the OWOD definition. In this section, we rethink the OWOD problem and summarize the basic principles to guide the benchmark construction. Then we propose two fair evaluation protocols specific to OWOD from the perspective of unknown classes.}




\begin{figure}[]
    \centering
    \includegraphics[scale = 0.24]{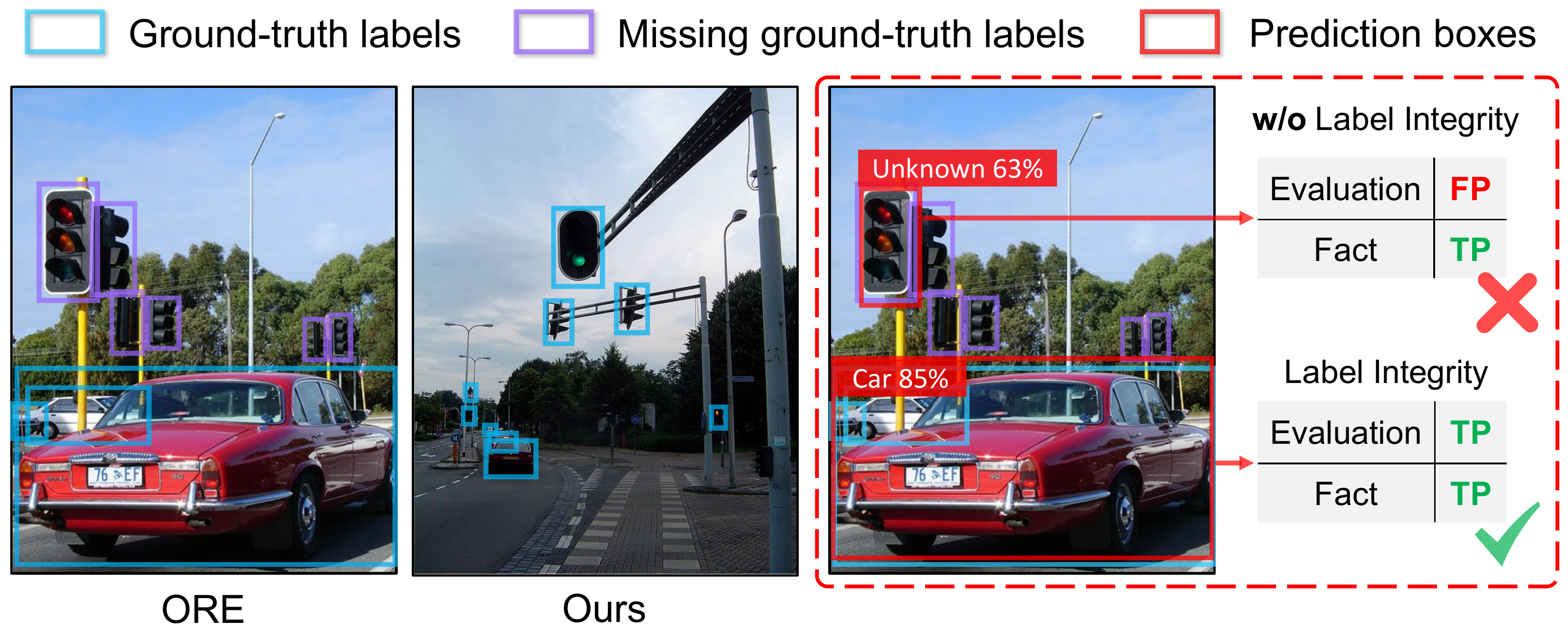}
    \caption{The influence of Label Integrity. Without complete annotations, the detected traffic lights will be mistakenly regarded as False Positive in benchmarks without Label Integrity. 
    }
    \label{img:test_img}
\end{figure}

\subsection{Benchmark Principles}\label{sec:principles}

{A high-quality benchmark could effectively help researchers to train, test, polish, and refine their models. Therefore, we summarize the benchmark principles to guide the OWOD benchmark construction, simulating realistic object detection in the open world. We consider an OWOD benchmark setting should meet the following principles:}



{\textbf{Class Openness:} This principle indicates that both known class set $\mathcal{K}$ and unknown class set $\mathcal{U}$ may exist during inference, where $\mathcal{K}\cap\mathcal{U}=\varnothing$. However, only known classes are labeled and used to train the detector.
}



{\textbf{Task Increment:} This principle states known classes are increasing in size, and thus the task is incrementally developed. Assuming the known set and unknown set in the $t$-th task are denoted as $\mathcal{K}_{t}$ and $\mathcal{U}_{t}.$ 
In task $t+1$, the unknown classes of interests, without loss of generality, termed as $\mathcal{U}_{t}^{(k)} \in \mathcal{U}_{t}$, are annotated and added in known set $\mathcal{K}_{t+1}=\mathcal{K}_{t} \cup \mathcal{U}_{t}^{(k)}$. The current unknown set $\mathcal{U}_{t+1}=\mathcal{U}_{t} \backslash \mathcal{U}_{t}^{(k)}$. This process continues until $\mathcal{U}_{T}=\varnothing$. }

\begin{figure}[]
    \centering
    \includegraphics[scale = 0.5]{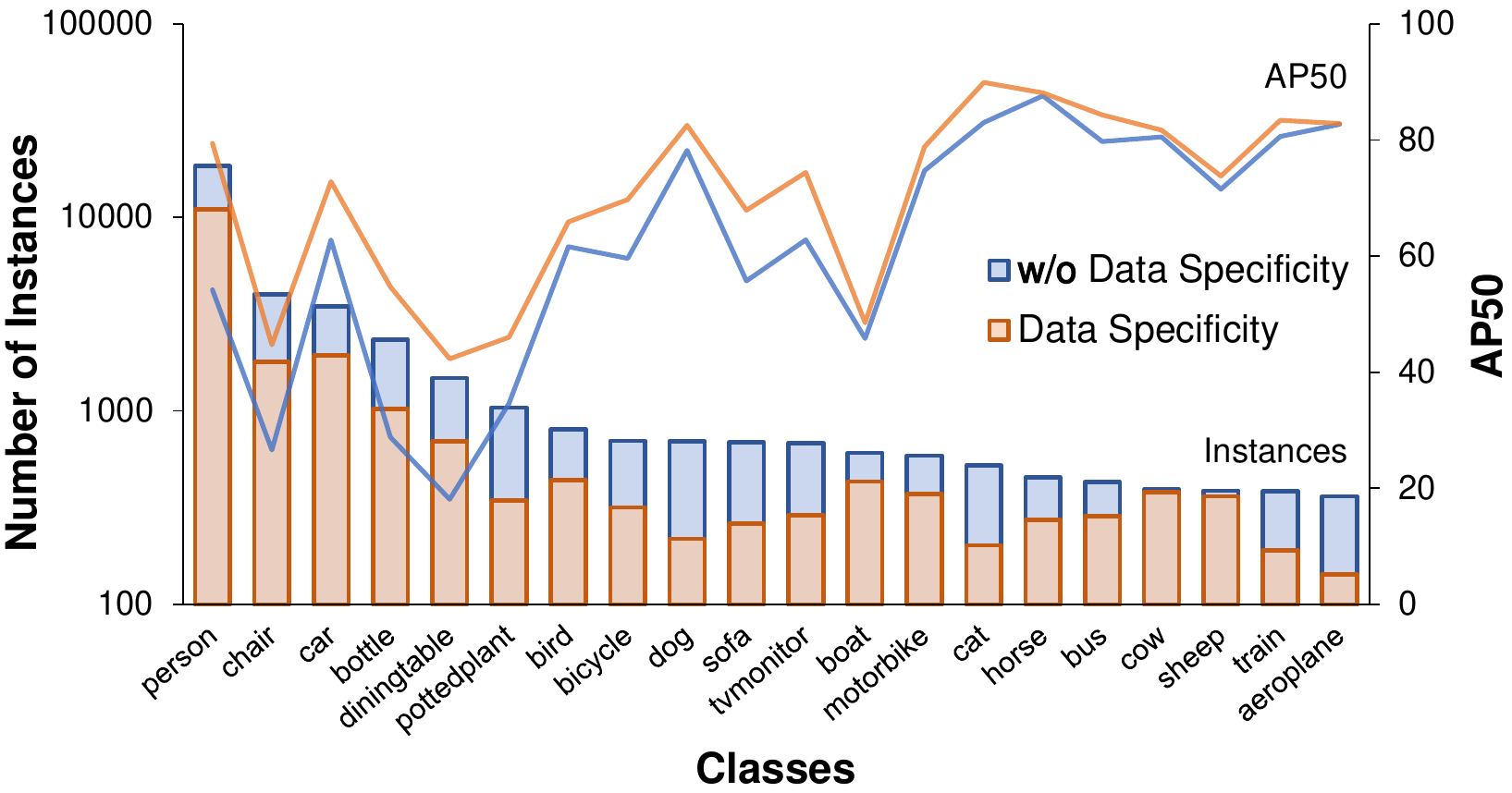}
    \caption{The influence of Data Specificity. Performance on benchmark without Data Specificity varied irregularly compared to the reasonable counterpart, indicating that it cannot reflect the actual performance of the model.
    }
    \label{img:data_specify}
    \vspace{-1em}
\end{figure}

{\textbf{Annotation Specificity}: This principle supports the different annotation strategies for the training, validation, and testing dataset. In the training or validating phase, only instances of known classes $\mathcal{K}$ are assigned labels $\mathbf{Y}=[\mathbf{L}, \mathbf{B}]$, where $\mathbf{L} \in \mathcal{K}$ indicates class labels while $\mathbf{B}$ is the location coordinates of its corresponding bounding box. During testing, except for labeling the instances of known classes, instances of unknown classes will only be assigned an ``unknown'' class label and their corresponding boxes coordinates. The previous work \cite{joseph2021towards} broke this principle and mistakenly used the fully-annotated validation set with unknown-class information involved to train an energy-based classifier, achieving misleading gains.}

\textbf{Label Integrity:} This principle points that the label information should be integrated during test for a fair evaluation. Based on this principle, we argue that some standard object detection benchmarks cannot be directly used as OWOD benchmarks. For instance, the PASCAL VOC dataset \cite{everingham2010pascal} only involves 20 annotated classes with other existing classes unlabeled, leading to unreasonable ``false positive" due to lack of ground-truth annotations, as shown in Figure \ref{img:test_img}. 
However, the previous work \cite{joseph2021towards} adopt both PASCAL VOC and COCO datasets during testing, which is confusing and ill-suited. 




{\textbf{Data Specificity:} This principle contains two implications. For one thing, there is no intersection of the training set, validation set, and testing set. For another, there should be no duplication inside of each dataset. For instance, the benchmark proposed by \cite{joseph2021towards} contains multiple similar testing images, which may affect the evaluation, such as exacerbation of the error rate, as shown in Figure \ref{img:data_specify}. }


{According to these principles, we argue that the benchmark proposed by ORE \cite{joseph2021towards} cannot be adopted for a fair evaluation. Therefore, we construct a new benchmark for the OWOD task.
We summarize the different benchmark settings of the previous work ORE \cite{joseph2021towards} and ours in Table \ref{tab:princple}. According to the Label Integrity principle, we adopt the relatively fully-annotated COCO \cite{lin2014microsoft} dataset, {and adopt the validation set of COCO with 48 incompletely labeled images removed as our test set}. We split the 80 classes of the COCO dataset into four tasks distinct by the semantic super-category and select data from the classes of each task as training dataset, consistent with the Task Increment principle. The testing dataset is chosen from all classes, satisfying the Class Openness principle. Repetitive data are removed to satisfy the Data Specificity. Finally, we assign labels according to Annotation Specificity, and take 1000 data away from the original training dataset as the validation set in each task. The concrete data division are listed in Table \ref{tab:data}. Following all the benchmark principles, our benchmark could fairly evaluate the OWOD performance.
}

\subsection{Metrics}\label{sec:metric}

{Eligible evaluation metrics quantify the performance of a model targeted for a specific task. However, previous metrics cannot comprehensively evaluate the OWOD performance. Therefore, in this section, we first point out the main challenges lying in OWOD, and then state the defects of previous metrics in evaluating the detection performance for these challenges. At the end of the section, we introduce our newly-designed evaluation metrics UDP and UDR, which could effectively make up for the deficiency.}
\begin{table}[]
\resizebox{\linewidth}{!}{
\begin{tabular}{c|ccccc}
\hline
    & \tabincell{c}{Class\\Openness} & \tabincell{c}{Task \\Increment} & \tabincell{c}{Annotation\\ Specificity} &  \tabincell{c}{Label\\Integrity}&\tabincell{c}{Data\\ Specificity} \\ \hline
ORE \cite{joseph2021towards} & $\surd$        & $\surd$                   &   ×                & ×                & ×                  \\
Ours & $\surd$        & $\surd$                  & $\surd$                 & $\surd$              & $\surd$                \\ \hline
\end{tabular}}
\caption{Comparison of the benchmark settings of \cite{joseph2021towards} and ours.}
\label{tab:princple}
\end{table}

\begin{table}[]
  \centering
    \resizebox{\linewidth}{!}{
    \begin{tabular}{c|cccc}
    \hline
          & Task 1 & Task 2 & Task 3 & Task 4 \\
    \hline
    \tabincell{c}{Semantic\\Split} & \tabincell{c}{Person\\Vehicle,Animal} & \tabincell{c}{Outdoor\\Accessories\\Appliance, Truck} & \tabincell{c}{Sports\\Food} & \tabincell{c}{Electronic\\Furniture\\Indoor, Kitchen} \\
    \hline
    Train & 91903  & 43684  & 37677  & 38446  \\
    \hline
    Val   & 1000  & 1000  & 1000  & 1000  \\
    \hline
    Test  & \multicolumn{4}{c}{4952} \\
    \hline
    \end{tabular}}%
  \caption{ The task composition in our new benchmark.}
  \label{tab:data}%
  \vspace{-1em}
\end{table}%

{According to the OWOD definition, there are three key challenges: 1) \textbf{Unknown Objectness:} distinguishing an unknown instance from the background \cite{dhamija2020overlooked}. 2) \textbf{Unknown Discrimination:} distinguishing an unknown instance from a similar known class \cite{dhamija2020overlooked}. 3) \textbf{Incremental Conflict:} balance between the learning of existing known classes and the newly-annotated known classes. 
Since the third one is similar to the idea of incremental learning \cite{rebuffi2017icarl,shmelkov2017incremental}, we mainly concentrate on the evaluation of the Unknown Objectness and Unknown Discrimination challenges.
}

{Since the first challenge only exists in the object detection task, it is totally understandable that previous open world learning approaches, including both recognition methods and the only detection method \cite{joseph2021towards}, simply focus on the second challenge, namely the discrimination between unknown and known classes. 
Previous open world learning usually adopted the standard mAP to measure the performance of classes, and also utilized WI \cite{dhamija2020overlooked} and Absolute
Open-Set Error (A-OSE) \cite{miller2018dropout} metrics to evaluate the impact caused by forcing the model to detect unknown classes.
A-OSE reports the number of unknown objects that get wrongly classified as any of the known classes, denoted as $\text{FP}_o$.
We would also like to point out that the WI metric was first proposed by an open set object detection work, and thus its computational formula \textit{CANNOT} be directly applied to OWOD problem, as \cite{joseph2021towards} mistakenly did.\footnote{But, \cite{joseph2021towards} correctly implement WI metric in their codes.} The computational formula of the WI metric and its connection to the A-OSE metric are given as follows:}

\begin{equation}
 \text{WI}=\frac{\text{FP}_o}{\text{TP}_k+\text{FP}_k}=\frac{\text{A-OSE}}{{\text{TP}_k+\text{FP}_k} }.  
\end{equation}
where $\text{TP}_k$ and $\text{FP}_k$ means the true positive proposals and false positive proposals of current known classes. More discussions will be found in the appendix.


Apart from these traditional metrics which simply focuses on the unknown discrimination challenge from the perspective of known classes, in this paper, we also put forward two new metrics focusing respectively on Unknown Objectness and Unknown Discrimination from the perspective of unknown classes, namely Unknown Detection Recall (UDR) and Unknown Detection Precision (UDP). UDR indicates the accurate localization rate of unknown classes, while UDP demonstrates the accurate classification rate of all localized unknown instances. Specifically, we term the true positive proposals and false negative proposals of unknown classes respectively as $\text{TP}_u$, $\text{FN}_u$, {and $\text{FN}_u^*$ indicates the number of ground-truth boxes recalled by misclassified predicted bounding boxes.} Therefore, the UDP and UDR could be calculated as:  
\begin{equation}
    \text{UDR}=\frac{\text{TP}_u+\text{FN}_u^*}{\text{TP}_u+\text{FN}_u},
\end{equation}
\begin{equation}
    \text{UDP}=\frac{\text{TP}_u}{\text{TP}_u+\text{FN}_u^*}.
\end{equation}

\begin{figure}[]
    \centering
    \includegraphics[scale = 0.22]{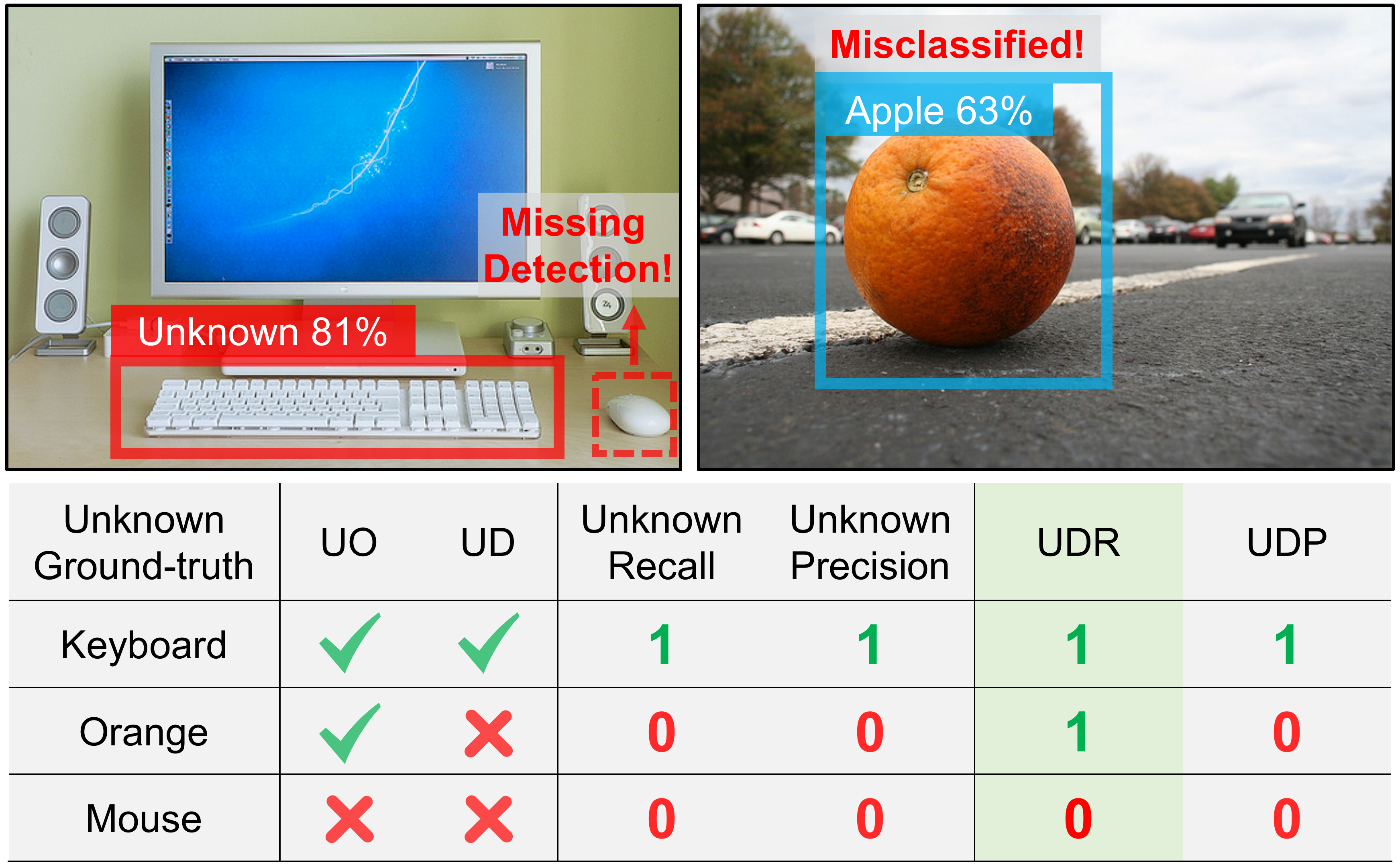}
    \caption{Rationality of the proposed metric. For different detection errors of unknown instances, the localization of ``Orange" is accurate, better than the missing detection of ``Mouse". However, classic metrics such as Recall and Precision equally treat these errors, while our metrics could distinguish them and better reflect the performance when facing Unknown Objectness (UO) and Unknown Discrimination (UD) challenges specific in OWOD.       
    }
    \label{img:metric}
    \vspace{-1em}
\end{figure}

\begin{figure*}[]
    \centering
    \includegraphics[scale = 0.48]{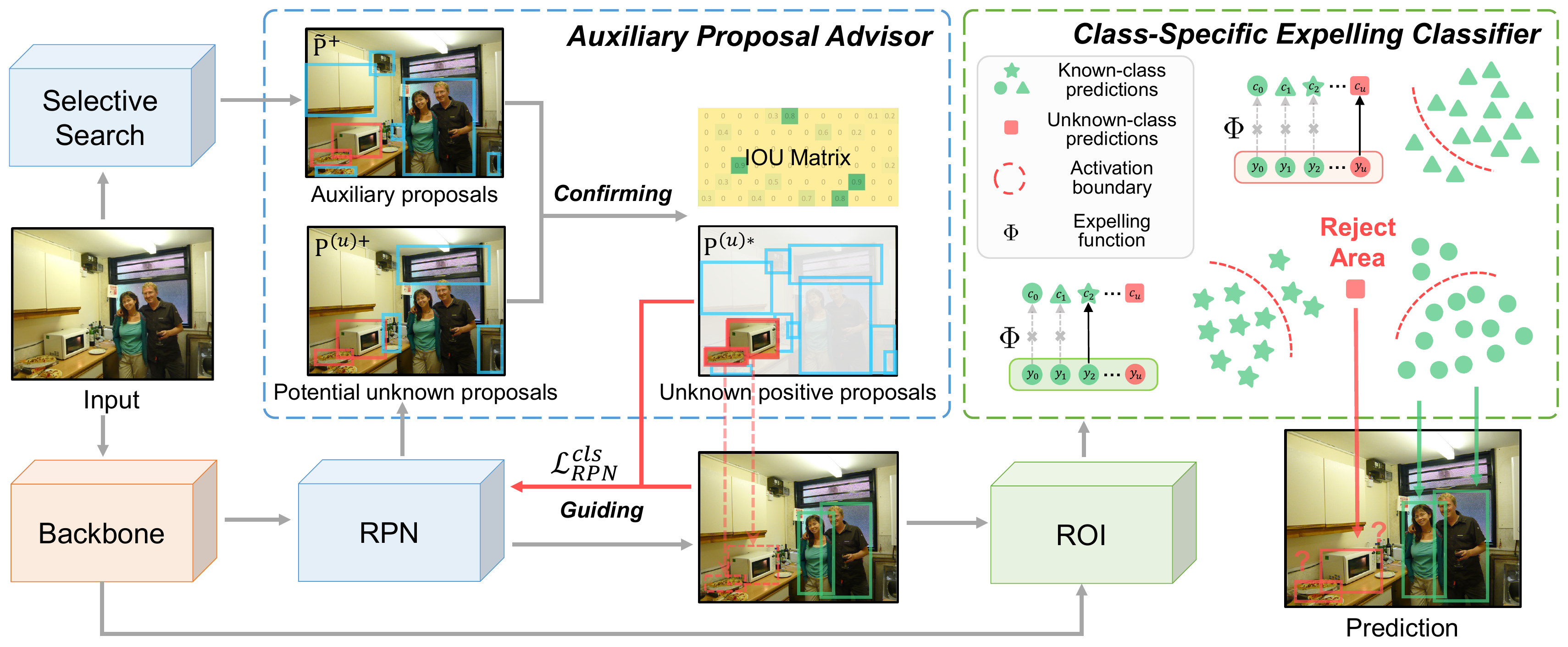}
    \caption{The architecture of our model. The Auxiliary Proposal Advisor confirms the potential unknown proposals generated by RPN and further guides the learning of RPN. The Class-specific Expelling Classifier calibrates the over-confident activation boundary for each class through the expelling function $\Phi$ to expel misclassified unknown instances.}
    \label{img:frame}
    \vspace{-1em}
\end{figure*}
{Similar to the classic metric Recall and Precision, UDR and UDP are complementary to each other and should be discussed together. Nevertheless, as the example in Figure \ref{img:metric}, the primary concern of UDP and UDR is the accurate localization rather than classification, which is a particular challenge in OWOD without annotations of unknown classes.  
Therefore, utilizing both the previous metrics and ours, we could comprehensively measure the performance of an OWOD model and make a fair comparison between different OWOD models.}

\section{Method}

{To alleviate the Unknown Objectness and Unknown Discrimination challenges, we further proposed an OWOD framework including an auxiliary Proposal ADvisor (PAD) and a Class-specific Expelling Classifier (CEC), as shown in Figure \ref{img:frame}. The non-parametric PAD could assist the RPN in identifying the unknown proposals, while the CEC adopts an expelling function to calibrate the over-confident known class activation boundary and re-allocate the predictions in a class-specific way. Specifically, we adopt the Faster R-CNN \cite{ren2015faster} as our base detector and insert the two novel modules into it. We will elaborate on the details of the proposed PAD and CEC in the following sections.}
{It is worth noting that, in each incremental task $t+1$, we first train the model including PAD and CEC with the newly-known classes, namely $\mathcal{U}_{t}^{(k)}$, and then fine-tune with the whole known classes $\mathcal{K}_{t+1}$. For clearer representation, the task notation $t$ is omitted in the following sections.
}

{\subsection{Auxiliary Proposal Advisor}} 

{Without label information of unknown classes, it is hard for the RPN to generate positive proposals for unknown classes confidently. Therefore, we deploy an auxiliary non-parametric Proposal ADvisor module (PAD) to assist the RPN in identifying the ambiguous proposals. This advisor is more than confirming the positive unknown proposals generated by RPN, but also guiding the learning of localizing unknown classes without supervision. Thus, the advisor and RPN collaboratively localize the positive proposals of unknown classes, which could pass the high-quality proposals for the successive detector and improve the final detection performance.} 

{\textbf{Confirming Proposals:} The RPN is trained to judge the objectness of a proposal through the supervision information of known classes. Therefore, it is much likely that the negative proposals with higher objectness scores are the positive unknown proposals that are misclassified as background without label information. Thus we denote these potential unknown positive proposals as ${\mathbf{P}}^{(u)+}$.}

{However, it is also possible that ${\mathbf{P}}^{(u)+}$ contains the authentic meaningless background. Without ground-truth labels, nothing can be sure of the unknown classes. That is to say, ${\mathbf{P}}^{(u)+}$, possibly containing both positive unknown proposals and meaningless background, are messy and uncertain. Their objectness scores $\mathbf{S}$ are also unreliable. }

{ Therefore, we need to introduce the advisor to confirm the objectness of ${\mathbf{P}}^{(u)+}$ and select more reliable positive proposals for further detection. The advisor produces the possible regions of objects, termed as $\widetilde{\mathbf{P}}^{+}$.  $\widetilde{\mathbf{P}}^{+}$ could be generated by any unsupervised object detection approaches. In our paper, we simply choose the classic non-parametric Selective Search \cite{uijlings2013selective}, which calculates these possible regions of objects through a hierarchical grouping of similar areas based on color, texture, size, and shape compatibility. Thereby the confirmation of the objectness of the $i$-th proposal of $\mathbf{P}^{(u)+}$  could be formulated as follows:
}
\begin{equation}
    \bar{\mathbf{S}_i}= \mathbf{S}_i \times \mathcal{I}\{\max \limits_{1\leq j \leq |\widetilde{\mathbf{P}}^{+}|} (\text{IOU}(\mathbf{P}^{(u)+}_i,\widetilde{\mathbf{P}}^{+}_j))>\theta\},
\end{equation}
{where $|\cdot|$ is the length of a set, while $\mathcal{I}(\cdot)$ is the Kronecker delta function that is equal to 1 when the input condition holds and 0 otherwise. {IOU (intersection-over-union), also known as Jaccard index, is the most commonly used metric for comparing the similarity between two arbitrary shapes\cite{everingham2010pascal}}. A larger IOU indicates that both the advisor and the RPN consider this region as a positive proposal of high objectness, which means the advisor confirms this proposal output by the RPN. $\theta$ is the threshold for the IOU score to filter the ambiguous proposals. We set a strict value as 0.7 to avoid introducing more messy boxes. Then we can select unknown proposals according to $\bar{\mathbf{S}_i}$.
}

{Through the confirmation for proposals, we could select positive unknown proposals in a more confident way and could filter out the messy proposals to save the computation and improve the further detection performance. }

{\textbf{Guiding Learning:} Since we could filter out the ambiguous proposals of  $\mathbf{P}^{(u)+}$ and obtain more accurate ones, these accurate proposals could be in turn used as the guidance to improve the localization ability of the RPN. Specifically, we modify the  
class label of these accurate proposals as ``foreground" and thus form the pseudo labels of these proposals. And then, we seek out their original anchors $\mathcal{A}^{(u)+}$ and remove them from the negative anchors set $\mathcal{A}^-$ to the positive anchor set $\mathcal{A}^+$. Finally, we fed the new anchor set into the class-agnostic classifier $f$ of RPN and calculate its loss function $\mathcal{L}_{RPN}^{cls}$ as follows:}

{
\begin{equation}\small
   \sum_{{\mathbf{a} \in \mathcal{A}^{+}\cup \mathcal{A}^{(u){+}}}} \text{BCE}(f(\mathbf{a}),\mathbf{1}) \ \ + \sum_{{\mathbf{a} \in{\mathcal{A}^{-}\setminus \mathcal{A}^{(u)+} }}}\text{BCE}(f(\mathbf{a}),\mathbf{0}),
\end{equation}
where BCE is the Binary Cross Entropy loss.}

\subsection{Class-Specific Expelling Classifier}
{Discriminative classifiers based on deep neural networks are often criticized for producing overconfidence predictions that cannot reflect the true probability of classification accuracy \cite{guo2017calibration}. In the open world setting, this problem will be more serious due to the lack of any explicit information of unknown classes, meaning that an instance of unknown classes would also be predicted as a known class with high confidence, namely Unknown Discrimination.  
}

{To overcome the Unknown Discrimination challenge, we propose a Class-specific Expelling Classifier to expel the confusing instances from the predicted known class and re-allocate their class predictions. We sufficiently utilize the annotation information of known classes to calibrate the over-confident activation boundary of each class. According to the refined activation boundary, a predicted bounding box could re-allocate its class predictions and determine its predicted classes. Specifically, an expelling indicator could be obtained through the expelling function $\Phi$ for each predicted box. 
Since each known class may have its class-specific activation area, we adaptively conduct the expelling operation inside each known class. After the refinement of the class activation boundary, if all classes determine to expel this box through their corresponding expelling indicator, we will predict it as the ``unknown" category. Otherwise, we will rank the confidence of the classes which allow its existence and choose the highest one as its class label.}

We first compute the class-specific expelling indicator through the expelling function $\Phi$.
As for the output prediction $\bar{\mathbf{Y}}=[\bar{\mathbf{L}}, \bar{\mathbf{B}}]$ of testing dataset, we denote $\bar{\mathbf{L}}^{c}_i$ as the probability whether the $i$-th sample belongs to the $c$-th known category. We could thereby compute the following expelling indicator of the $c$-th category for the $i$-th sample:
\begin{equation}\small
    \Phi(\bar{\mathbf{L}}^{c}_i)=\bar{\mathbf{L}}^{c}_i-\alpha \times \frac{1}{M}\sum_{{j}}^{|\widetilde{\mathbf{B}}|}\sum_{{k}}^{|\mathbf{B}^{c}|}[\mathcal{I}(\text{IOU}(\widetilde{\mathbf{B}}_j,\mathbf{B}^{c}_k)>\varphi)\times \widetilde{\mathbf{L}}^{c}_i],
\end{equation}
{where {$\widetilde{\mathbf{L}}=[\widetilde{\mathbf{L}},\widetilde{\mathbf{B}}]$ is the output predictions of the training images, $\mathbf{Y}=[\mathbf{L}, \mathbf{B}]$ is the corresponding ground-truth labels.} $M=\sum_j^{|\widetilde{\mathbf{B}}|}\sum_k^{|\mathbf{B}^{c}|}[\mathcal{I}(\text{IOU}(\widetilde{\mathbf{B}}_j,\mathbf{B}^{c}_k)>\varphi)]$ counts the number of samples satisfying the condition of the Kronecker delta function. $\varphi$ is the IOU threshold to choose the possible bounding boxes of the $c$-th category, while $\alpha$ is the hyper-parameter to adjust the expelling degree.}

Then, we could re-allocate the known class predictions for a predicted bounding box:
\begin{equation}
    \bar{\mathbf{L}}^{{c}'}_i = \mathcal{I}(\Phi(\bar{\mathbf{L}}^{c}_i)>0) \times  \bar{\mathbf{L}}^{c}_i.
\end{equation}
{Moreover, if for all known classes, we set the predicted logits of the ``unknown" class as 1, namely $\bar{\mathbf{L}}^{(|\mathcal{K}|+1)'}_i$=1. After the re-allocation, we could determine its class from the calibrated prediction.}

\begin{table*}[htbp]
  \renewcommand\arraystretch{1.5}
  \centering
  \setlength\tabcolsep{2.2pt}
    \resizebox{\linewidth}{!}{
    \begin{tabular}{c|c|c|ccc|c|ccc|ccc|c|ccc|ccc|ccc}
    \hline
    Task IDs & \multicolumn{5}{c|}{Task 1}   & \multicolumn{7}{c|}{Task 2}                   & \multicolumn{7}{c|}{Task 3}                   & \multicolumn{3}{c}{Task 4} \\
    \hline
    \multirow{2}[5]{*}{} & WI    & mAP (↑) & UR & UDR   & UDP   & WI    & \multicolumn{3}{c|}{mAP (↑)} & UR & UDR   & UDP   & WI    & \multicolumn{3}{c|}{mAP (↑)} & UR & UDR   & UDP   & \multicolumn{3}{c}{mAP (↑)} \\
    \cline{3-3}\cline{8-10}\cline{15-17}\cline{21-23}          & (↓)   & current & (↑)  & (↑)   & (↑)   & (↓)   & previous & current & both  & (↑)  & (↑)   & (↑)   & (↓)   & previous & current & both  & (↑)  & (↑)   & (↑)   & previous & current & both \\
    \hline \rowcolor[rgb]{0.851, 0.851, 0.851}
    Oracle & 0.04144  & 61.00  & 68.75 & 71.01  & 96.82  & 0.02863  & 58.96  & 47.66  & 53.31  & 67.46 & 69.75  & 96.73  & 0.02097  & 53.38  & 38.54  & 48.43  & 68.49 & 72.35  & 94.67  & 49.43  & 38.14  & 46.60  \\
    \hline
    Faster-RCNN & 0.05332  & \textbf{60.53}  & 0.00 & 20.48  & 0.00  & 0.05047  & 0.75  & \textbf{37.84}  & 19.30  & 0.00 & 24.52  & 0.00  & 0.03312  & 1.90  & \textbf{30.53}  & 11.44  & 0.00 & 24.92  & 0.00  & 0.56  & \textbf{31.48}  & 8.29  \\
    \hline
    Faster-RCNN & \multicolumn{5}{c|}{\multirow{2}[2]{*}{Not applicable as incremental}} & \multirow{2}[2]{*}{0.03957} & \multirow{2}[2]{*}{53.80} & \multirow{2}[2]{*}{37.07} & \multirow{2}[2]{*}{45.43}& \multirow{2}[2]{*}{0.00} & \multirow{2}[2]{*}{23.62} & \multirow{2}[2]{*}{0.00} & \multirow{2}[2]{*}{0.02928} & \multirow{2}[2]{*}{42.96} & \multirow{2}[2]{*}{26.56} & \multirow{2}[2]{*}{37.49} & \multirow{2}[2]{*}{0.00}& \multirow{2}[2]{*}{36.20} & \multirow{2}[2]{*}{0.00} & \multirow{2}[2]{*}{37.83} & \multirow{2}[2]{*}{28.73} & \multirow{2}[2]{*}{35.55} \\
    +Finetuning & \multicolumn{5}{c|}{}         &       &       &       &       &       &       &       &       &       &       &       &       &       &       &  \\
    \hline
    ORE*  & 0.05409  & 60.30  & 3.46 & 20.67  & 16.71  & 0.04054  & 54.08  & 37.79  & \textbf{45.94}  & 3.29 & 23.68  & 13.87  & 0.02763  & \textbf{43.30}  & 26.18  & \textbf{37.59}  & 3.78 & 33.55  & 11.25  & 37.96  & 28.89  & \textbf{35.69}  \\
    \hline
    ORE   & 0.04575  & 59.56  & 7.76 & 20.39  & 38.04  & 0.03769  & 53.61  & 37.53  & 45.57  & 5.66 & 23.44  & 24.15  & 0.02570  & 41.78  & 25.91  & 36.49  & 9.92 & 33.17  & 29.91  & 37.96  & 28.89  & \textbf{35.69} \\
    \hline
    Ours  & \textbf{0.04493}  & 59.70  &  \textbf{9.08} & \textbf{21.20}  & \textbf{42.87}  & \textbf{0.03306}  & \textbf{54.11}  & 37.26  & 45.64  & \textbf{9.89} &\textbf{24.62}  & \textbf{40.18}  & \textbf{0.02412}  & 43.06  & 24.64  & \textbf{37.59} & \textbf{11.36}  & \textbf{36.83}  & \textbf{33.81}  &\textbf{37.99}  & 28.66  & {35.66}  \\
    \hline
    \end{tabular}}%
  \caption{OWOD results on our new benchmark.  The ``previous", ``current" and ``both" in ``mAP''  represent mAP of previously known classes, newly-added known classes and all known classes of the task, while ``UR" means Recall of unknown classes. The \textbf{bold} indicates the best performance. (↑) means higher is better, and (↓) means lower is better. Our approach consistently keeps the performance on known classes and outperforms baseline methods by large margins on the unknown classes. }
  \label{tab:com}%
  \vspace{-1em}
\end{table*}%


\section{Experiments}
In this section, we compare our model with the state-of-the-art detection models to demonstrate its effectiveness, according to both the widely-used evaluation metrics (WI, mAP, {Recall}) and our new protocols (UDR, UDP). Moreover, we meticulously analyze each component of the proposed framework and visualize the detection performance. More experiments with other evaluation metrics can be found in the appendix.

\subsection{Implementation Details}

{
We train our networks on 8 Tesla V100. Our model is based on the standard Faster R-CNN \cite{ren2015faster} with a ResNet-50 \cite{he2016deep} backbone. The proposed benchmark contains 4 incremental tasks. In each task, we set the batch size as 8 and run the SGD optimizer with the initial learning rate of 0.01. The momentum is set as 0.9 and the weight decay is set as 0.0001. In task 1, the overall iterations are 90k for sufficient training. While the training iterations of the newly-known classes in each incremental task are respectively 52k, 40k, and 41k, with the following fine-tuning iterations of 4k, 2.5k, 3k. We respectively select top-50 potential unknown proposals from RPN and top-50 auxiliary proposals from the auxiliary proposal advisor. $\theta$, $\varphi$ and $\alpha$ are empirically set as 0.7, 0.9 and 0.5.
}





\begin{figure}[]
    \centering
    \includegraphics[scale = 0.17]{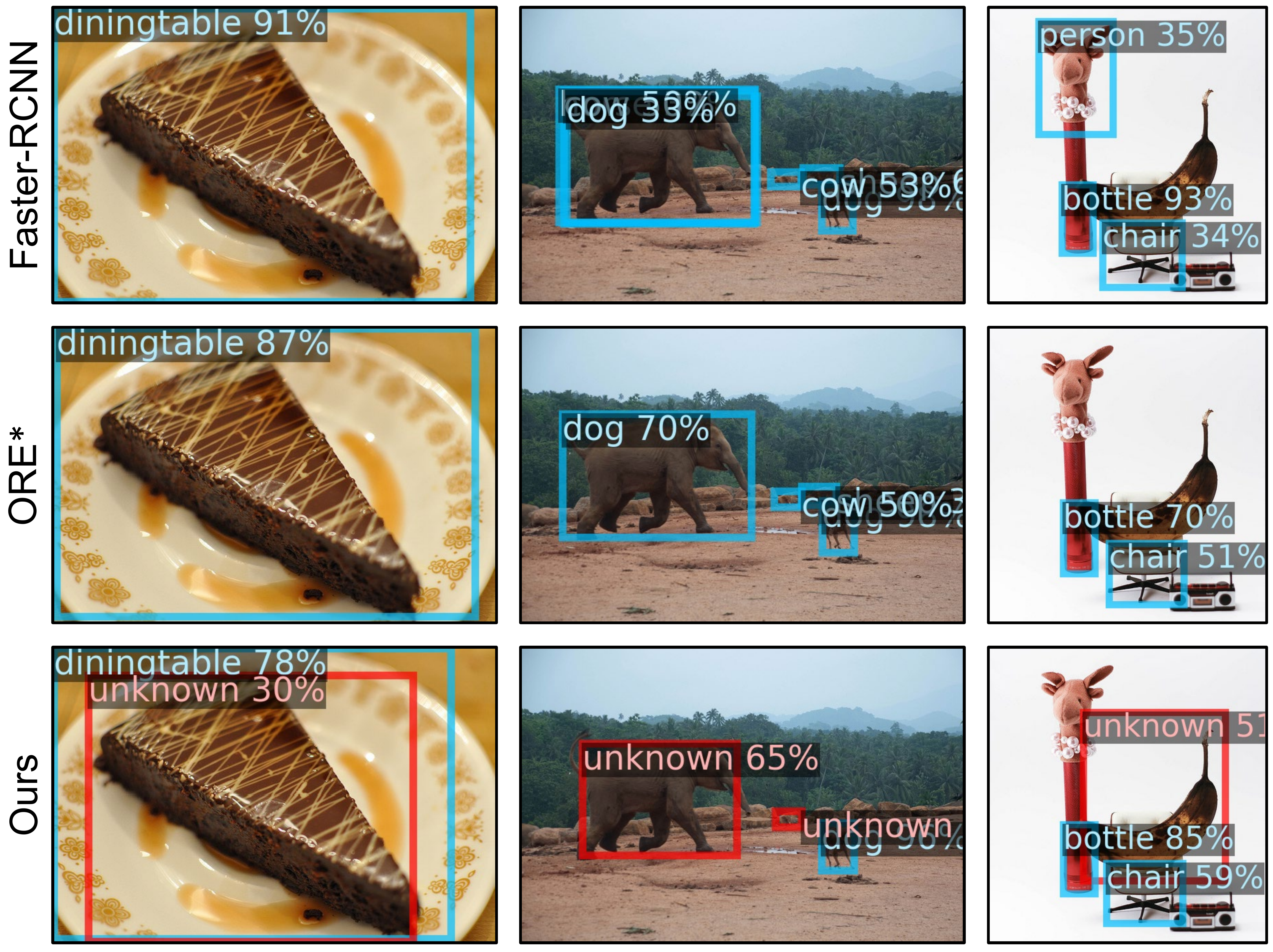}
    \caption{Visualization results of Faster-RCNN, ORE* and Ours.}
    \label{img:result}
    \vspace{-1em}
\end{figure}

\subsection{Compare with the State-of-the-art}
{We compared the proposed model with state-of-the-art models including 1) the base detector Faster-RCNN \cite{ren2015faster}, 2) Faster-RCNN + finetuning in each incremental task, 3) ORE* model which is the ORE model without its unknown annotation-leakage EBUI module (energy-based unknown identification), 4) ORE model \cite{joseph2021towards}, and 5) Oracle detector which is trained with full labels of known and unknown classes. The Oracle can be regarded as the upper limit of OWOD performance, for reference only in our experiments.}

{Table \ref{tab:com} lists the performance of the proposed method and other detectors. We can observe that our model achieves superior performance in almost all the settings. From the WI performance, the advantageous performance fully demonstrates that our model has minimal impact on known classes detection. Comparing the mAP of known classes among different tasks, we can observe that, our model performs well on both previously known classes and currently known classes, proving its incremental learning ability. In contrast, without incremental designs, although the original Faster-RCNN obtain the best performance on the currently known classes, it cannot work on the previous ones. Since there is unknown annotation-leakage in the original ORE, the performance of ORE is legitimately better than the fair ORE*. However, the proposed model can outperform ORE and ORE*, proving its effectiveness.  
}

{Moreover, our model brings obvious gains according to UR, UDP and UDR, illustrating that it could better distinguish the unknown classes from the background and avoid confusion with known classes in the meanwhile. 
The even more strong localization ability for unknown classes, according to UDR, states that the obvious improvement of UDP mainly gains from the authentic enhancement of the classification performance between known classes and unknown classes, instead of false improvement through reducing the number of localized boxes.} We also visualize the comparison results in Figure \ref{img:result}. It can be seen that Faster-RCNN and ORE* show good performance on known classes with unknown instances missed or misclassified, while our method can accurately localize and classify unknown instances as ``unknown".

{
From these analyses, we can conclude that our OWOD model shows strong detection performance on both known classes and unknown classes with incremental learning ability, which is more applicable in the real world. 
}

\subsection{Analysis and Discussion}
{In this section, we first verify the validity of each component, and then conduct a sensitivity analysis of the CEC. We mainly report the performance in task 1.}

{\textbf{Ablation for Components:} Table \ref{tab:component} respectively lists the results of 1) the base detector, 2) the base detector with PAD, 3) the base detector with CEC, and 4) the whole model. It can be observed each module contributes to our model, while combined with both modules the model provides maximum performance gains. Specifically, the improvement of UDR is completely brought from PAD, meaning that PAD contributes to the accurate localization and alleviates the Unknown Objectness. Moreover, both PAD and CEC benefit the UDP performance, declaring they contribute to Unknown Discrimination as well. Furthermore, when deploying the two modules, PAD provides accurate unknown proposals for CEC, achieving peak performance.
} 

{\textbf{Analysis for PAD:} Table \ref{tab:advisor} shows the performance of different implementations of PAD.  
The model with an original RPN cannot precisely distinguish unknown classes from the known classes, leading to poor UDP performance. 
The confirmation operation can help provide accurate unknown proposals for further classification while still maintaining the performance of known classes. Moreover, with the guiding operation, the proposed model can gradually improve the confidence of unknown instances and screen out the messy unknown proposals, yielding the best performance.}

\begin{figure}[]
    \centering
    \includegraphics[scale = 0.2]{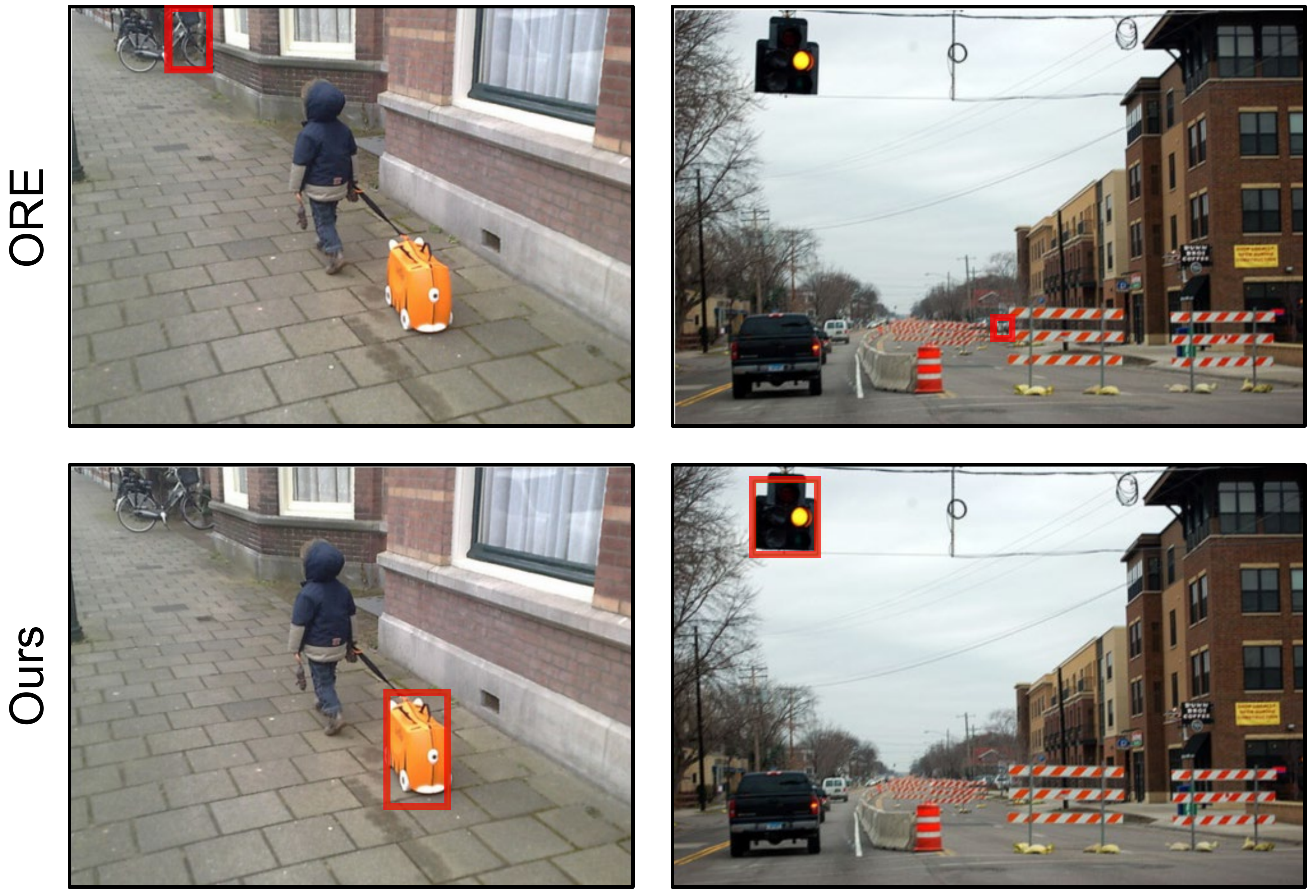}
    \caption{Comparison of selected unknown proposals from ORE model and our PAD.}
    \label{img:unk_vis}
    \vspace{-1em}
\end{figure}

{Figure \ref{img:unk_vis} shows some cases of the positive unknown proposals generated by the original RPN and the RPN with the PAD module. When only using the RPN trained by known labels, the selected boxes from negative proposals of high objectness score may contain authentic background. With the assistance of the PAD, the auxiliary proposals can help confirm the proposals and guide the learning of the original RPN to screen out the background.
}

\begin{table}[]
\tiny
\centering
\resizebox{\linewidth}{!}{
\begin{tabular}{c|cccc}
\Xhline{0.2pt}
Method                                             & WI(↓)   & mAP(↑)                                 & UDR(↑)                                 & UDP(↑)                                 \\ \Xhline{0.2pt}
Baseline                                                             & 0.05345  & 60.49                                                  & 20.57 & 16.43                               \\
Baseline+PAD                       & 0.05247 & \textbf{60.51} & \textbf{21.20} & 22.91 \\
{Baseline+CEC} &   0.05252  	 	    & 	60.25                             & 20.67                           & 21.42                                 \\
Ours                                               & \textbf{0.04690} & 60.10                                  & \textbf{21.20}                                 & \textbf{36.74}     \\ \Xhline{0.2pt}
\end{tabular} }
\caption{Ablation study for our key components. }
\label{tab:component}
\end{table}

\begin{table}[]
\setlength\tabcolsep{3.5pt}
\resizebox{\linewidth}{!}{
\begin{tabular}{l|ccc|cccc}
\hline
ID &                                    RPN & confirming  & guiding & WI(↓)   & mAP(↑)                                & UDR(↑)                                 & UDP(↑)                                 \\ \hline
 1&$\surd$   &         &         & 0.05345 & 60.49                                  & {20.57} & 16.43 \\
     2&                                   $\surd$   & $\surd$       &         & 0.05370 & 60.38                                  & 20.41                                  & 22.53                                  \\
3& $\surd$  & $\surd$       & $\surd$       & \textbf{0.05276} & \textbf{60.51} & \textbf{21.20} & \textbf{22.91} \\ \hline
\end{tabular}
}
\caption{Analysis for different implementations in our PAD.}
\label{tab:advisor}
\end{table}

\begin{figure}[]
    \centering
     \includegraphics[scale = 0.48]{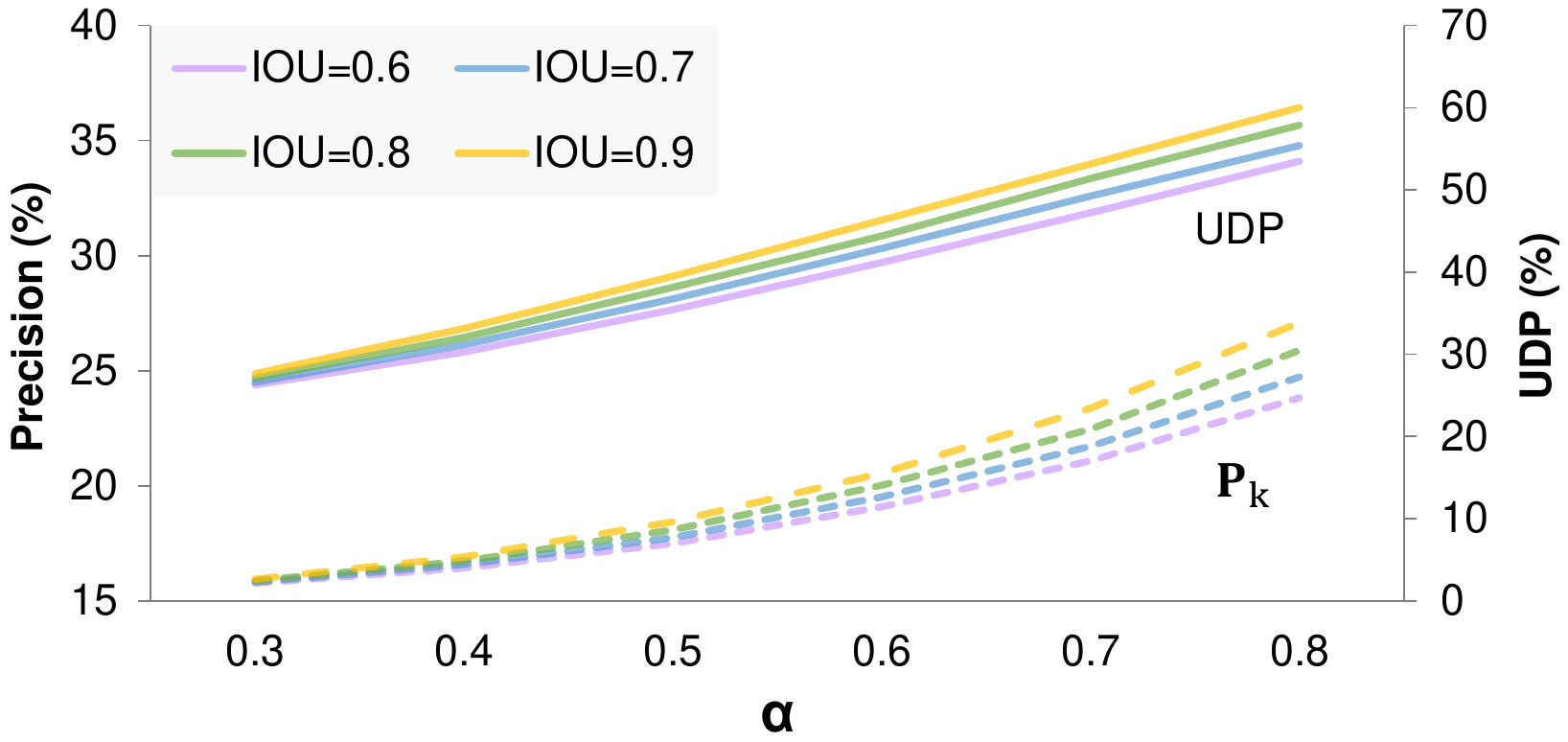}
     \caption{Sensitivity analysis on class-specific expelling classifier. $\textbf{P}_{\text{k}}$ indicates the precision of known classes.}
     \label{img:postprocessing}
     \vspace{-1em}
 \end{figure}

{\textbf{Sensitivity Analysis on CEC:} {Figure \ref{img:postprocessing} shows the precision change of both unknown and known classes as the hyper-parameters $\alpha$ and $\varphi$ in the CEC vary on testing dataset.} Consistent with our intuition, strict IOU will improve the accurate classification for unknown classes and maintain the performance on known classes. With the increment of $\alpha$, {we can observe that both the precision of unknown and known classes are improving, indicating the CEC could accurately expel the unknown instances from a known class to ensure the classification precision. However, it will harm the recall performance and thus influence the overall mAP. More analysis will be found in the appendix.} 

Following the Annotation Specificity principle, we should choose hyper-parameter with the validation set without annotation. In our experiments, we empirically choose strict IOU threshold $\varphi$ as 0.9, and select $\alpha=0.5$ satisfying that mAP performance of known classes drops within $1\%$ on validation set. We can observe that, the expected $\alpha$ might be 0.6 on the testing dataset with both less mAP drop and much UDP improvement.}
  

\section{Conclusion}
{
In this work, we found the defects of the only open world object detection (OWOD) work and revisited this task. We proposed five fundamental principles to guide the OWOD benchmark construction and two fair evaluation protocols UDP and UDR specific to OWOD task. 
Moreover, we presented a new OWOD framework including an auxiliary Proposal ADvisor and a Class-specific Expelling Classifier. 
Comprehensive experiments conducted on our fair benchmark demonstrated the effectiveness of our method and also validated the rationality of our metrics. We will release our benchmark and code, and hope this work could contribute to the community and facilitate further insights. 
}


\newpage
{\small

\bibliographystyle{ieee_fullname}
}

\end{document}